\documentclass[conference]{IEEEtran}
\IEEEoverridecommandlockouts
\usepackage{cite}
\usepackage{amsmath,amssymb,amsfonts}
\usepackage{algorithmic}
\usepackage{graphicx}
\usepackage{textcomp}
\usepackage{xcolor}
\def\BibTeX{{\rm B\kern-.05em{\sc i\kern-.025em b}\kern-.08em
    T\kern-.1667em\lower.7ex\hbox{E}\kern-.125emX}}

\newcommand{\ignore}[1]{}

\begin{document}

\title{Learning to Score}

\author{\IEEEauthorblockN{Yogev Kriger}
\IEEEauthorblockA{\textit{Efi Arazi School of Computer Science} \\
\textit{Reichman University}\\
yogev.kr@gmail.com}
\and
\IEEEauthorblockN{Shai Fine}
\IEEEauthorblockA{\textit{Efi Arazi School of Computer Science} \\
\textit{Reichman University}\\
shai.fine@runi.ac.il}
}


\maketitle

\begin{abstract}
Common machine learning settings range from supervised tasks,
where accurately labeled data is accessible, through semi-supervised and weakly-supervised tasks, where target labels are scant or noisy, to unsupervised tasks where labels are unobtainable. 
In this paper we study a scenario where 
the target labels are not available but additional related information is at hand. This information, referred to as Side Information, is either correlated with the unknown labels or imposes constraints on the feature space. We formulate the problem as an ensemble of three semantic components: representation learning, side information and metric learning. The proposed scoring model is advantageous for multiple use-cases. For example, in the healthcare domain it can be used to create a severity score for diseases where the symptoms are known but the criteria for the disease progression are not well defined.
We demonstrate the utility of the suggested scoring system on well-known benchmark data-sets and bio-medical patient records.

\end{abstract}


\section{Introduction}
Various data-driven methods have been developed over the years; supervised methods were devised for cases where ample labeled data is available. Semi-supervised methods were developed for cases where labels are well-defined but the labeled data is less available and difficult to acquire. 
Weak supervision has been suggested to handle cases where labels are either noisy, imprecise, or incomplete (i.e. weakly labeled data).
Unsupervised methods were developed for situations where it is difficult or impossible to obtain labeled data. 

We introduce a model that aims to solve problems for which we do not observe any label, but labels do exist. 
The goal of the model is to create a score that would represent the true, unseen labels. This is achieved by utilizing characteristics of the instances and additional data, referred to as side information. Following the learning process, the model produces a score that is representative of the unseen labels.

The proposed framework showcases its versatility by accommodating a wide spectrum of scenarios, ranging from cases where the side information offers minimal insight about the label, to cases where it is richly informative. This spectrum encapsulates two extremes - an unsupervised task, where the side information bears no correlation to the label, and a supervised task, where the side information is, in fact, the label itself.

It is applicable for many problems in various fields: Consider the emergence of a novel, previously unidentified disease. The Medical community lacks the requisite knowledge to define a disease-severity score, not knowing which factors contribute to the disease prognosis.  However, the experts could provide some reasonable assumptions. For example, observing a decrease in blood saturation is positively correlated with worse prognosis. In this case, saturation could be used as side information and combined with additional clinical records of the patients as model input, could be used to create a disease-severity score. Another possible application is scoring marketing leads using life-time value (an estimate of the revenue customers will generate) as the side information. The model can also be used for quality assurance, to score the yield of an assembly line using, as a side information, lot level production factors and SLAs (service-level agreement) along the manufacturing process. In the educational sector, the model can leverage related academic performance metrics as side information to enrich the understanding of students' performance in other domains, paving the way for a more holistic and individualized approach to student assessment.

We created such a generic model by combining several machine learning techniques. In a nutshell, our approach can be described as representation learning constrained by the side information. The model receives the side information in two possible forms: information that is related to the labels and information related to the underlying structure of the data that is incorporated using metric learning. We used a generative modeling approach to construct a latent space but strived to form a representation that is as discriminative as possible. 

The rest of the paper is organized as follows - next, we describe a few related works. In section~\ref{sec:Background} we provide details of the fundamental concepts we used for the development of the model. In section~\ref{sec:methodology} we introduce and thoroughly explain the various components of the scoring model. In section~\ref{sec:experiments} we perform multiple ablation studies and demonstrate the contribution of each component to the model. Furthermore, we show the ability of the model to handle various scenarios, ranging from supervised to unsupervised. 
We then turn to illustrate how the scoring model can address real-life setting, by applying it to learn the well known UPDRS severity score for Parkinson Disease patients, using the motoric UPDRS as side information.
We conclude in section~\ref{sec:summary} with some insights and suggestions for further research.
\subsection{Related work}
To some extent, score learning is a novel task. In what follows, we present the most related tasks and models, and highlight the similarities and differences to score learning.
\subsubsection{Variational Information Bottleneck for Semi-Supervised Classiﬁcation}

In their paper, Voloshynovskiy et al.~\cite{VST:20} construct a new formulation of Information Bottleneck (IB) for semi-supervised classiﬁcation and use a variational decomposition to convert it into a practically tractable setup with learnable parameters. Additionally, they suggest using learnable priors on the latent space of classiﬁers together with different regularizes. Our multi-task architecture can be viewed as dimensionality reduction with regularizes, similar to the architecture suggested in their work. Both works share concepts and components such as IB, reconstruction and classification space constraints. However, the setup is different - in our model there is no direct knowledge of the target or access to the target labels. Rather, knowledge of the target is provided indirectly, in the form of side information, and constraints over the feature space, which in turn are being utilized by the metric learning component.

\subsubsection{Learning using privileged information}
Learning using privileged information (LUPI)~\cite{vv:09} is a machine learning paradigm where models are trained using both standard input data and additional "privileged" information. This privileged information can include data that is not available during test time, such as detailed annotations of training data, and can be used to improve model performance on a wide range of tasks. For example, when the objective is to classify biopsy images to two categories - images that contain a cancerous tumor and images that do not. When the images are accompanied by pathologists reports, describing the images in a high level holistic manner, they could be used as "privileged" information to find a good classification rule in the pixel space. The privileged information can be considered as a form of side information. However the LUPI setting differs from ours, since on one hand we don't assume any access to the target labels, and on the other hand we may allow access to the side information during test time as well.  

\subsubsection{Weak supervision}
Weak supervision~\cite{rhr:19} is a technique used in machine learning where models are trained using weakly labeled data, instead of a fully labeled data. Weakly labeled data refers to data that is not exhaustively labeled, such as data with labels that are noisy, imprecise, or incomplete. This approach can help overcome the limitations of traditional supervised learning that relies on large amounts of fully and accurately labeled data, which is often difficult and expensive to acquire. The weakly labeled data can be considered as a form of side information, making weak supervision a special case of our suggested setting.

\section{Essentials}
\label{sec:Background}
Our proposed model integrates the following components: 
representation learning, learning with side information and metric learning.

\subsection{Representation learning}
Representation learning is the process of learning a set of features or a representation of the input data that is useful for a specific task.

Structure-focused methods like Autoencoders use a neural network that effectively seeks to learn the intrinsic structure of the data in an unsupervised manner. It learns a representation typically by reducing the dimensionality of the data and reconstructing it back to the original dimensions while trying to keep the output data as close as possible to its original input. 
Of a particular interest is the Variational Autoencoder (VAE)~\cite{kw:13}. VAE utilizes a probabilistic framework to learn complex data distributions. From representation learning perspective, VAE is a structure-focused method that learns a stochastic mapping to a (latent) representation. The model consists of an encoder and a decoder network, where the encoder maps the input data to a latent representation, and the decoder generates a reconstruction of the input data from the latent representation. The VAE model is trained by minimizing a loss function that measures the difference between the input data and its reconstruction, as well as a regularization term of the latent representation to encourage it to be a smooth and continuous distribution.

\subsection{Information Bottleneck}
Information bottleneck (IB) is a concept in information theory that refers to the balance between preserving the relevant information in a signal or data set while minimizing the amount of unnecessary or redundant information. It was first introduced by Tishby et al.~\cite{T:00}. The IB can be thought of as a limiting factor in the flow of information, as it controls how much information is retained or discarded based on its relevance to the task at hand. Alemi et al.~\cite{M:16} proposed to use variational inference to construct a lower bound on the IB objective and call it variational information bottleneck (VIB). This allows to use deep neural networks to parameterize the distributions, and thus to handle high-dimensional, continuous data and avoid the previous restrictions to the discrete or Gaussian cases. 
In the original paper introducing VIB, the target is to maximize the information about the label. In this study, we incorporated VIB for maximizing information about the side information.

\subsection{Learning with side information}
\begin{figure}[htbp]
\centerline{\includegraphics[scale=0.35]{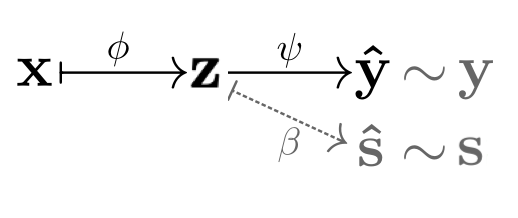}}
\caption{The pattern
assumes that the target function $f~=~\mathcal{\psi}~\circ~\mathcal{\phi}$ and the related
function $g~=~\mathcal{\beta}~\circ~\mathcal{\phi}$ share $\mathcal{\phi}$, and therefore have the same intermediate representation $z = \mathcal{\phi}(x)$. By training the representation to predict both $y$ using $\mathcal{\psi}$, and $s$ using the auxiliary function $\mathcal{\beta} : z \rightarrow s$, we incorporate the assumption that related tasks share intermediate representations. \cite{jrs:15}}
\label{fig:multi-task}
\end{figure}

Learning with side information refers to the use of additional data or knowledge to aid the learning process. This can be viewed from an information theory perspective as the incorporation of side information into the learning process to improve the efficiency and effectiveness of the learning algorithm. Side information can come in various forms. For example,  prior knowledge about the distribution of the data or about related tasks, additional data sources that may provide context or features for the learning task. One of the patterns to incorporate side information, which is most relevant to this study, is the Multi-Task pattern illustrated in Figure~\ref{fig:multi-task} and formalized by Jonschkowski et al.~\cite{jrs:15} as an output of an auxiliary function that shares part of the computations with the main function.

\subsection{Metric learning}

The technique involves learning of a distance metric or a measure of the similarities between data points
in order to perform tasks such as classification or clustering~\cite{bellet2015metric}. 
One popular method for metric learning is through the use of triplet loss, which involves training a model on a set of triplets of data points where each triplet consists of a query point, a positive example, and a negative example. In unsupervised tasks positive and negative examples can be inferred using self supervision. The model is trained to minimize the distance between the query point and the positive example, while simultaneously maximizing the distance between the query point and the negative example. This allows the model to learn a distance metric that can accurately reflect the underlying similarity or dissimilarity of the data, allowing it to make more accurate predictions or groupings. Triplet loss is often used in applications such as image or text classification, where the distance between data points may not be easily defined using traditional methods (c.f.~\cite{v:18}, \cite{hean:14}). In their paper, Ren et al.~\cite{RJ:21} propose a novel metric learning loss for distribution learning, the Jensen-Shannon (JS) triplet loss. This method reﬂects the similarity between two distributions. However, it should be noted that the triangle inequality doesn't hold for JS, hence it is not a true metric. In this study we chose to use the square root of the Jensen-Shannon divergence , which is indeed a metric~\cite{ES:03}, as a distance for the metric learning.

\section{The Scoring Model}
\label{sec:methodology}
\begin{figure}[tbp]
\centerline{\includegraphics[scale=0.4]{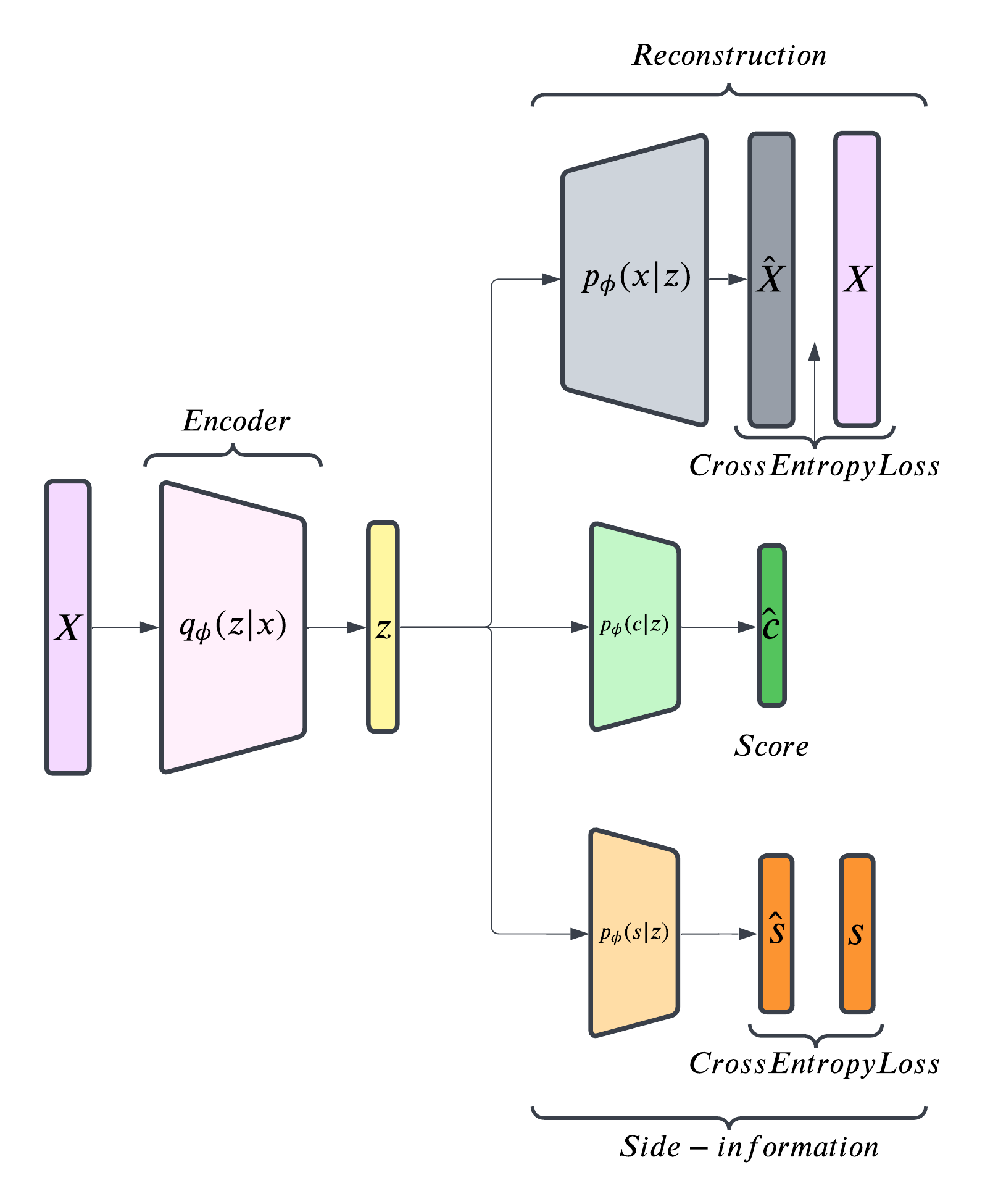}}
\caption{The model's architecture - composed of three parts: reconstruction, side information and score inference.}
\label{fig:architecture}
\end{figure}

Our proposed model is assembled from three semantic components: representation learning, side information and metric learning. Their synergistic integration results with a predictive scoring model. The objective function was developed such that its minimization would optimize the learning of the components, and consequently optimize the learning of the entire model. Representation learning is performed by mapping the instances in the feature space into a lower dimension latent space using reconstruction and triplet loss, thus preserving the topological local and global structure and capturing the discriminative information of the data. Score learning is performed by mapping the vectors from the latent space to a numeric or categorical score. All components of the model are depicted in figure~\ref{fig:architecture} and are thoroughly described in the following sections.

\subsection{Model Input and Encoding}\label{input}
The model can receive various  data types as input, denoted as $X$ in figure \ref{fig:architecture}. The input data propagate through a stochastic encoder that maps it to a low-dimensional latent representation. The output of the encoder is represented by $q(Z \mid x_n)$.
The encoder architecture can be adjusted to support various data types - for tabular data, a simple multi-layer perceptron (MLP) is often used, while for images, more advanced DNNs such as VGGNet or ResNet are frequently utilized.
A regularization term that affects the latent representation is the KL divergence between the posterior distribution $q(z \mid x)$ and a prior distribution $p(z)$, such as a standard normal distribution. It encourages the latent representation to be similar to the prior distribution. By minimizing the KL divergence, the model is able to learn a latent representation that is "smooth" and approximates the prior distribution. We added the following KL divergence term to the loss.

\begin{equation}
\begin{aligned} 
 \mathcal{L}_{KL} = D_{KL} \left( q (Z \mid x_n ) \parallel p \left(Z \right) \right) 
\end{aligned}
\end{equation}

As can be seen in figure \ref{fig:architecture}, the encoder is shared and affected by the three branches. 

\subsection{Reconstruction}\label{VAE}
In order to learn meaningful and coherent latent representations of the data, we utilized a VAE. Given an input data point $x$, the VAE consists of two main components: an encoder, described in the section above, and a decoder $p(x \mid z)$ that maps the latent representation back to the original high-dimensional space and is shown in figure~\ref{fig:architecture}. The VAE is trained by minimizing the negative log likelihood of the data, also known as the reconstruction loss $L(x,x')$, where $x'$ is the reconstructed data point obtained by passing the input $x$ through the encoder network and then back through the networks. 
We added the decoder term of the VAE to the loss in the following manner:
\begin{equation}
\begin{aligned} 
\mathcal{L}_{R} = \frac{1}{N} \sum_{n=1}^{N} - \log p\left(x_n \mid x_n'\right)
\end{aligned}
\end{equation}

\subsection{Side information}
We incorporated the side information, denoted by $S$ in figure~\ref{fig:architecture}, in a multi task pattern. Specifically, inspired by VIB we maximised the mutual information between the latent space $Z$ and the side information $S$:
\ignore{
The original objective function of VIB is defined as follows:

\begin{equation}
R_{IB}(\theta)=I(Z,Y;\theta)-\beta I(Z,x;\theta)
\end{equation}
where Z is the latent space, Y is the label, $\beta$ is a Lagrange multiplier and $x$ is the input. We adapted their approach in order to propagate information between the latent space and the side information:
}

\begin{equation}
R_{IB}(\theta)=I(Z,S;\theta)-\beta I(Z,x;\theta)
\end{equation}
where $\beta$ is a Lagrange multiplier.

In accordance with the Information Bottleneck principle, we assumed that the joint distribution $p(X,S,Z)$ factors as follows:

\begin{align}
\begin{split}
p(X, S, Z) &= p(Z \mid X, S )p(S  \mid X)p(X) \\
&= p(Z \mid X)p(S  \mid X)p(X)
\label{t}
\end{split}
\end{align}
\label{side_info}
This assumption is valid because the variables can be represented as the following Markov chain: $S \rightarrow X \rightarrow Z$. Hence $S$ and $Z$ are statistically independent given $X$ and we can assume $p(Z \mid X,S) = p(Z \mid X)$. Following this assumption, the representation $Z$ cannot depend directly on the side information $S$, which is the case of our multi task pattern. We can derive the following lower bound:

\begin{equation}
\mathcal{J}_{IB}=\frac{1}{N} \sum_{n=1}^{N}\left[-\log p(s_n \mid x_n)]+\beta D_{KL}[q(Z \mid x_n),q(Z)\right]
\end{equation}

Similar to the VAE in section \ref{VAE}, the KL divergence term of the Information Bottleneck is shared between the tasks, as described in section \ref{input}. Therefore, the contribution of the Information Bottleneck to the loss is the following:

\begin{equation}
\mathcal{L}_{SI}=\frac{1}{N} \sum_{n=1}^{N}-\log p(s_n \mid x_n)
\end{equation}

\subsection{Score prediction}
As the score is unknown, we try to recover discriminative information from the latent representation. We do it by maximizing the mutual information between $Z$ and the score $C$, i.e. $I(Z,C)$.
It is easy to show that $I(Z,C)$ can be decomposed as follows:

\begin{equation}
I(Z,C)=H(C)-H(C \mid Z)
\end{equation}
Where $H(C)$ is the score entropy and $H(C \mid Z)$ is the conditional entropy of the score given the instances in the latent space. Maximizing $I(Z,C)$ leads to minimizing $H(C \mid Z)$. Lowering the conditional entropy $H(C \mid Z)$ increases the discriminative capability that the latent space $Z$ has over the score $C$.
The loss can be constructed as follows:

\vspace{-0.5cm}
\begin{equation}
\begin{aligned} 
&\mathcal{L}_{score}= - \sum_{c \in C} [p(c) \log p(c)] \\
&+ \frac{1}{N}\sum_{n=1}^{N}[p \left(f\left(x_n \right)\right) p\left(c_n \mid f\left(x_n\right)\right) \log p\left(c_n \mid f\left(x_n\right)\right)]
\end{aligned}
\end{equation}
where $f\left(x_n \right)$ is the embedding of $x_n$ in the latent space. 

\subsection{Metric learning and the Square-Root-Jensen-Shannon Triplet Loss}

Triplet loss is a popular loss function used in deep learning to train models for metric learning tasks. It is defined as the differences of distance between the anchor (query) point, a positive example, and a negative example in the embedding space, with a margin $m$ imposed to ensure that the positive example is closer to the anchor point than the negative example:

\vspace{-0.2cm}
\begin{equation}
\mathcal{L} = \max \left (0, d \left( f \left(a \right) - f \left( p\right)\right) - d\left(f\left(a\right) - f\left(n\right)\right) + m\right)
\end{equation}
where $f(a)$, $f(p)$, and $f(n)$ are the embeddings of the anchor point, positive example, and negative example, respectively, and $d(\cdot,\cdot)$ denotes the Euclidean distance for a pair of embeddings. The triplet loss aims to minimize the distance between the anchor point and the positive example while simultaneously maximizing the distance between the anchor point and the negative example. 
In our proposed model, we assume that embeddings are representing distributions in the latent space.
Euclidean distance is less appropriate to handle distributions.
Hence, we adapted the square root of the Jensen–Shannon $(\sqrt{JS})$, which is metric for distributions (c.f. \cite{ES:03}), as the distance measure of the triplet loss:

\vspace{-0.2cm}
\begin{equation}
\label{eq:jsd}
\sqrt{JS(P \parallel Q)}= \sqrt{\frac{1}{2}D_{KL}(P \parallel M) + \frac{1}{2}D_{KL}(Q \parallel M)}
\end{equation}
where $M = \frac{1}{2}(P + Q)$. 

As can be seen in equation \ref{eq:jsd} the Jensen–Shannon divergence includes the calculation of two KL divergences. 
\ignore{
Suppose that we have two multivariate normal distributions with means $\mu_{0},\mu_{1}$ and covariance matrices $\Sigma _{0},\Sigma _{1}$ that have the same dimension $k$ -  then the KL divergence between the distributions has a closed form solution.
However, in equation \ref{eq:jsd} the KL divergence is performed on $P$ and $M$, where $M$ is an average of the two distributions but it is not necessarily a normal distribution, causing the absence of a closed form solution. On account of that, we replaced the average of the two distributions with their weighted geometric mean $G_\lambda(x,y)=x^{1-\lambda}y^\lambda$ for $\lambda \in [0,1]$ and we got skew-geometric Jensen-Shannon, $JS^{G_\lambda}$. Using the property that the weighted product of exponential family distributions stays in the exponential family (c.f.~\cite{nfgv:09}) the following closed from is derived~\cite{dsn:20}:
}
In equation \ref{eq:jsd}, $M$ is an average of the two Gaussian distributions, $P$ and $Q$, but $M$ itself is not a normal distribution, causing the absence of a closed form solution to both KL divergences. On account of that, we replaced the average of the two distributions with their weighted geometric mean $G_\lambda(x,y)=x^{1-\lambda}y^\lambda$ for $\lambda \in [0,1]$ and we got skew-geometric Jensen-Shannon, $JS^{G_\lambda}$. Using the property that the weighted product of exponential family distributions stays in the exponential family (c.f.~\cite{nfgv:09}), the following closed from is derived~\cite{dsn:20}:

\vspace{-0.2cm}
\begin{align}
\begin{split}
&JS^{G_\lambda} (\mathcal{N}_1 \parallel \mathcal{N}_2) \\
&= (1-\lambda)D_{KL}(\mathcal{N}_1 \parallel \mathcal{N}_\lambda)+\lambda D_{KL}(\mathcal{N}_2 \parallel \mathcal{N}_\lambda)\\ 
&= \frac{1}{2}\left(tr\left(\Sigma^{-1}_\lambda((1-\lambda)\Sigma_1+\lambda\Sigma_2)\right)
+ \log\left(\frac{|\Sigma_\lambda|}{|\Sigma_1|^{1-\lambda}|\Sigma_2|^\lambda}\right) \right. \\
&~~~~~~~~~~~~+ (1-\lambda)(\mu_\lambda - \mu_1)^T\Sigma_\lambda^{-1}(\mu_\lambda - \mu_1)\\
&~~~~~~~~~~~~+\lambda(\mu_\lambda - \mu_2)^T\Sigma_\lambda^{-1}(\mu_\lambda - \mu_2)-n\Bigg)
\end{split}
\end{align}

\ignore{
where $\mathcal{N}_1$, $\mathcal{N}_2$ and $\mathcal{N}_\lambda$ are all multivariate Gaussians, $\mathcal{N}_1(\mu_1, \Sigma_1)$, $\mathcal{N}_1(\mu_1, \Sigma_1)$ and $\mathcal{N}_1(\mu_\lambda, \Sigma_\lambda)$ where $\Sigma_\lambda = \left((1-\lambda)\Sigma_1^{-1}+\lambda \Sigma_2^{-1}\right)^{-1}$ and $\mu_\lambda~=~\left((1-\lambda)\Sigma_1^{-1}\mu_1+\lambda \Sigma_2^{-1}\mu_2 \right)$
}
where $\mathcal{N}_1(\mu_1, \Sigma_1)$, $\mathcal{N}_2(\mu_2, \Sigma_2)$ and $\mathcal{N}_\lambda(\mu_\lambda, \Sigma_\lambda)$ are all multivariate Gaussians, $\mu_\lambda~=~\left((1-\lambda)\Sigma_1^{-1}\mu_1+\lambda \Sigma_2^{-1}\mu_2 \right)$ and $\Sigma_\lambda = \left((1-\lambda)\Sigma_1^{-1}+\lambda \Sigma_2^{-1}\right)^{-1}$.

Using the skew-geometric Square-Root-Jensen-Shannon, denote as $\sqrt{JS^{G_\lambda}}$ we compute the triplet loss as follows:
\begin{equation}
\begin{aligned} \mathcal{L}_{triplet} = \sum_{(x_\lambda, x_p, x_n)}^{} \max \Big(0, \sqrt{JS^{G_\lambda}(f(x_a) \parallel f(x_p))} \\
-\sqrt{JS^{G_\lambda}(f(x_a) \parallel f(x_n))} + m \Big) 
 \end{aligned}
 \end{equation}
By that, we achieve a triplet loss that is variance aware, symmetric and confers robustness to the training process.
The triplets (anchor points, positive examples and negative examples) are defined in various ways and depend on both the setup and side information. In case that classes of the instances are available, the anchors and positive examples can be sampled from the same class and negative samples can be sampled from another class. In other cases, self supervision techniques can be utilized. 

\subsection{Optimization criteria}

By combining all components we get the following optimization criteria:

\begin{align}
\begin{split}
\mathcal{L} = &\mathcal{L}_{R+KL+triplet+SI+score}=\\
&\alpha \cdot \frac{1}{N} \sum_{n=1}^{N} [- \log p(x_n \mid x_n')]\\
+ &\beta \cdot \frac{1}{N} \sum_{n=1}^{N} [D_{KL} (q(Z \mid x_n) \parallel p(Z))] \\
+ &\gamma \cdot \sum_{(x_a, x_p, x_n)} \max \left(0, \sqrt{JS^{G_a}(f(X_a) \parallel f(X_p))} \right.\\
&~~~~~~~~~~~~~~~~~~~~~~~~~~\left.-\sqrt{JS^{G_a}(f(X_a) \parallel f(X_n))} + m \right) \\
+ &\delta \cdot \frac{1}{N} \sum_{n=1}^{N}[-\log q(s_n \mid x_n)] \\ 
+ &\zeta \cdot \left( - \sum_{c \in C} [p(c) \log p(c)] \right.\\ 
&\left.+ \frac{1}{N}\sum_{n=1}^{N}[p(f(x_n)) p(c_n \mid f(x_n)) \log p(c_n \mid f(x_n))] \right)
\end{split}
\end{align}

Note that the loss is comprised of five components, where both the triplet loss term and the KL divergence term are regularizations of the latent space $Z$. The other three components represent specific losses of the three branches depicted in figure~\ref{fig:architecture}: the reconstruction, the side information and the score.
\section{Experiments}
\label{sec:experiments}
Our model is designed to introduce a novel score for unlabeled tasks. Due to the nature of this problem we can compare results under specific and limited settings. We chose tasks for which there are both known scores and relevant side information. We do not use the labels in the training of our model, but rather solely to evaluate its performance.     

We present empirical results of our model for two known and publicly available datasets - MNIST \cite{D:12} and Parkinsons Telemonitoring Data Set \cite{AMA:09}. In each sub-section we begin by describing the dataset and the experimental setup, followed by the results of our model evaluation. We used Weights \& Biases \cite{wandb} for experiment tracking, performance visualizations, and to extract insights.
 
To assess the performance of our model we compared it against the performance of several baseline models. The results show that our model achieves comparable results to equivalent supervised models. Additionally, we conducted a series of ablation studies to understand the impact of various model components on performance. These results are presented in the following sections.

\subsection{Ablation studies using MNIST}
The MNIST dataset is a database of gray-scale handwritten digits. It consists of 60,000 training images and 10,000 testing images, all of which are 28x28 pixels in size and have been pre-processed to fit into a fixed-size matrix. 
We will use this dataset to obtain a better understanding on how the different components of the model behave.

The architecture of the encoder, which maps the instances from the feature space to the latent space, is a simple convolutional neural network (CNN) with two convolutional layers and three fully connected layers. More advanced encoder architectures can be used as necessary. 

To better understand each part of the model, we will examine the impact at the latent space while tuning the weight for each part.
For all the following figures, the latent space is higher than two dimensions, therefore in order to plot it, we used PCA to reduce dimensionality.

When only the reconstruction part is active, our model is essentially a classic autoencoder, or specifically a variational autoencoder.

\begin{figure}[htbp]
\centerline{\includegraphics[scale=0.15]{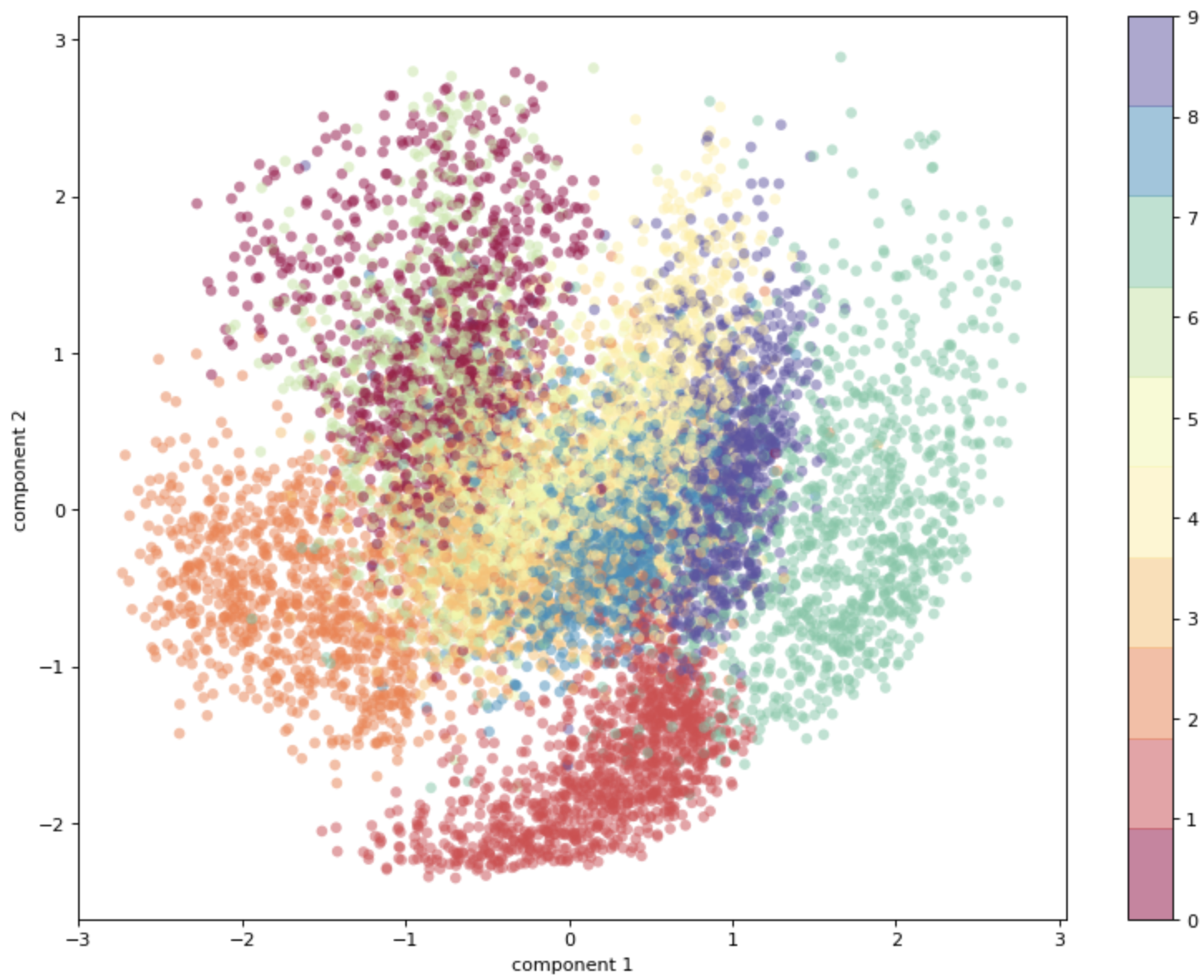}}
\caption{VAE - Latent space visualization of the test images.}
\label{fig:ls:vae}
\end{figure}

It can be seen in figure \ref{fig:ls:vae} how the model was able to learn the intrinsic structure of the data and how the different clusters are separable in the latent space. 

\begin{figure}[htbp]
\centerline{\includegraphics[scale=0.15]{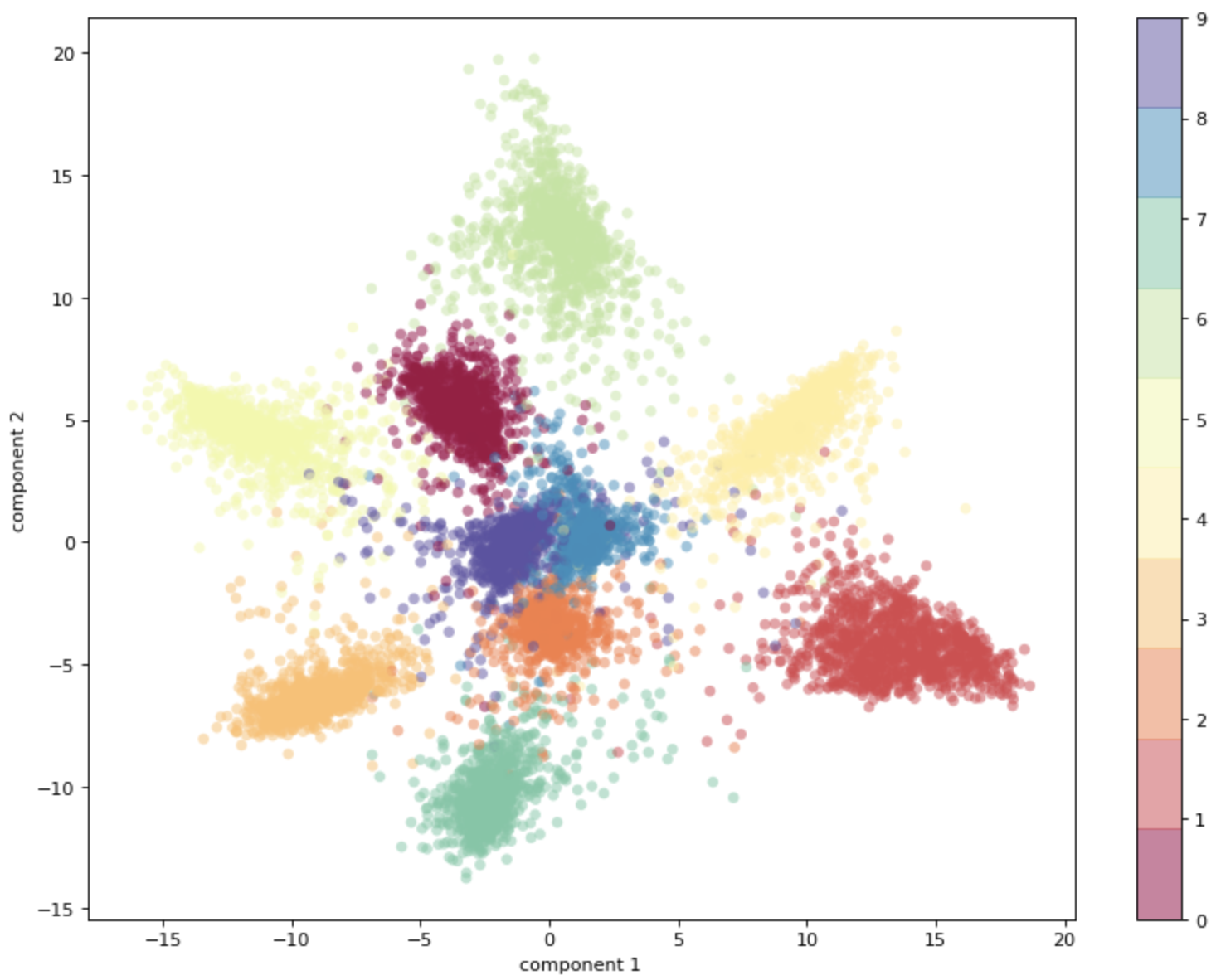}}
\caption{Triplet loss - Latent space visualization of the test images.}
\label{fig:ls:triplet}
\end{figure}

In figure \ref{fig:ls:triplet} only the triplet loss component is active. It can be seen that clusters are more separable while the variance of the latent space is high.

\begin{figure}[htbp]
\centerline{\includegraphics[scale=0.15]{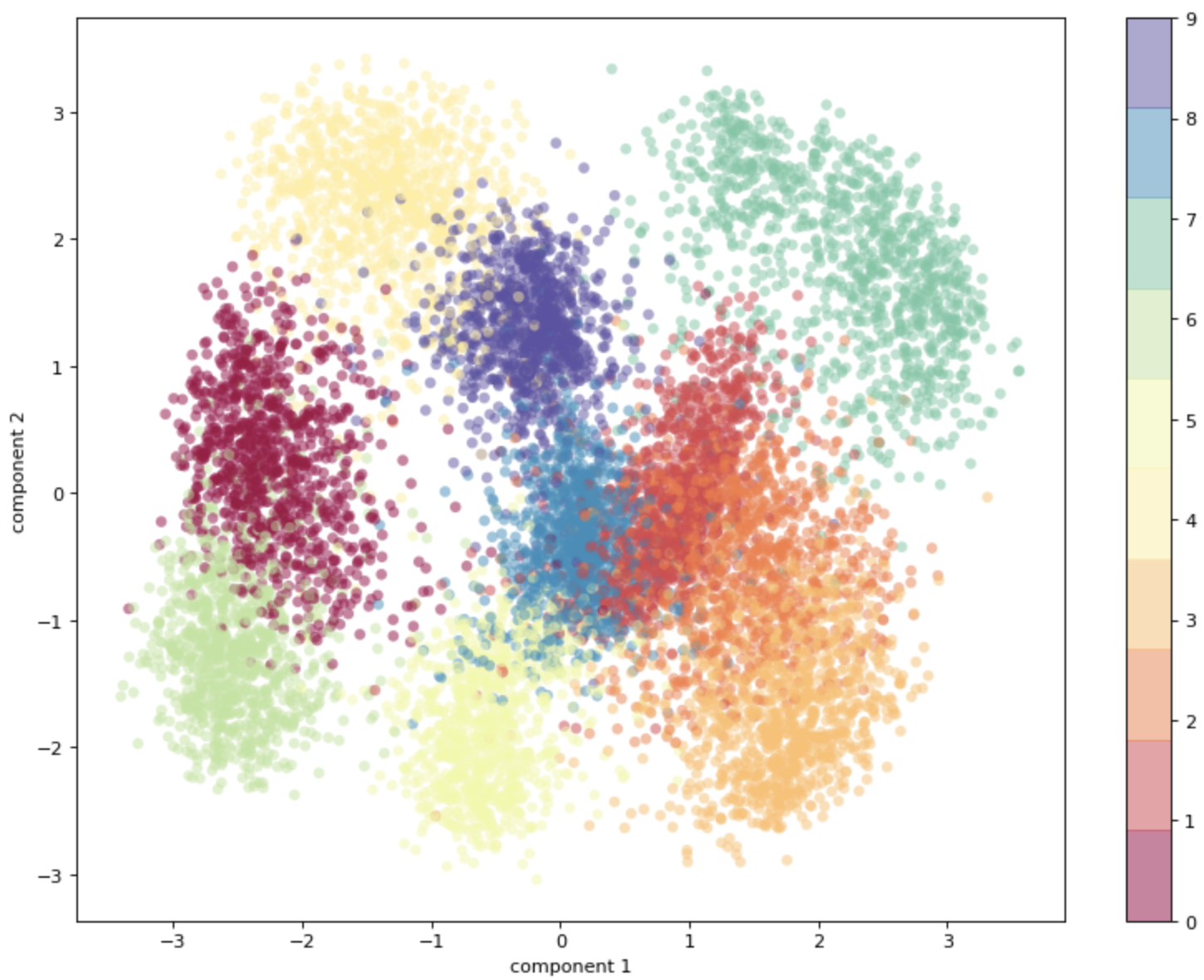}}
\caption{VAE \& Triplet loss - Latent space visualization of the test images.}
\label{fig:ls:vae_triplet}
\end{figure}

In figure \ref{fig:ls:vae_triplet} we can see the results of running the VAE part together with the triplet loss. We can see that by including these two parts we are able to achieve both a good separation between clusters and a continuous latent space with low variance. We will show how it contributes to the prediction of the score.

Next we would like to understand the impact of the side information on the latent space. 
For this aim we will share different variations of side information and the corresponding latent spaces.

\begin{figure}[htbp]
\centerline{\includegraphics[scale=0.15]{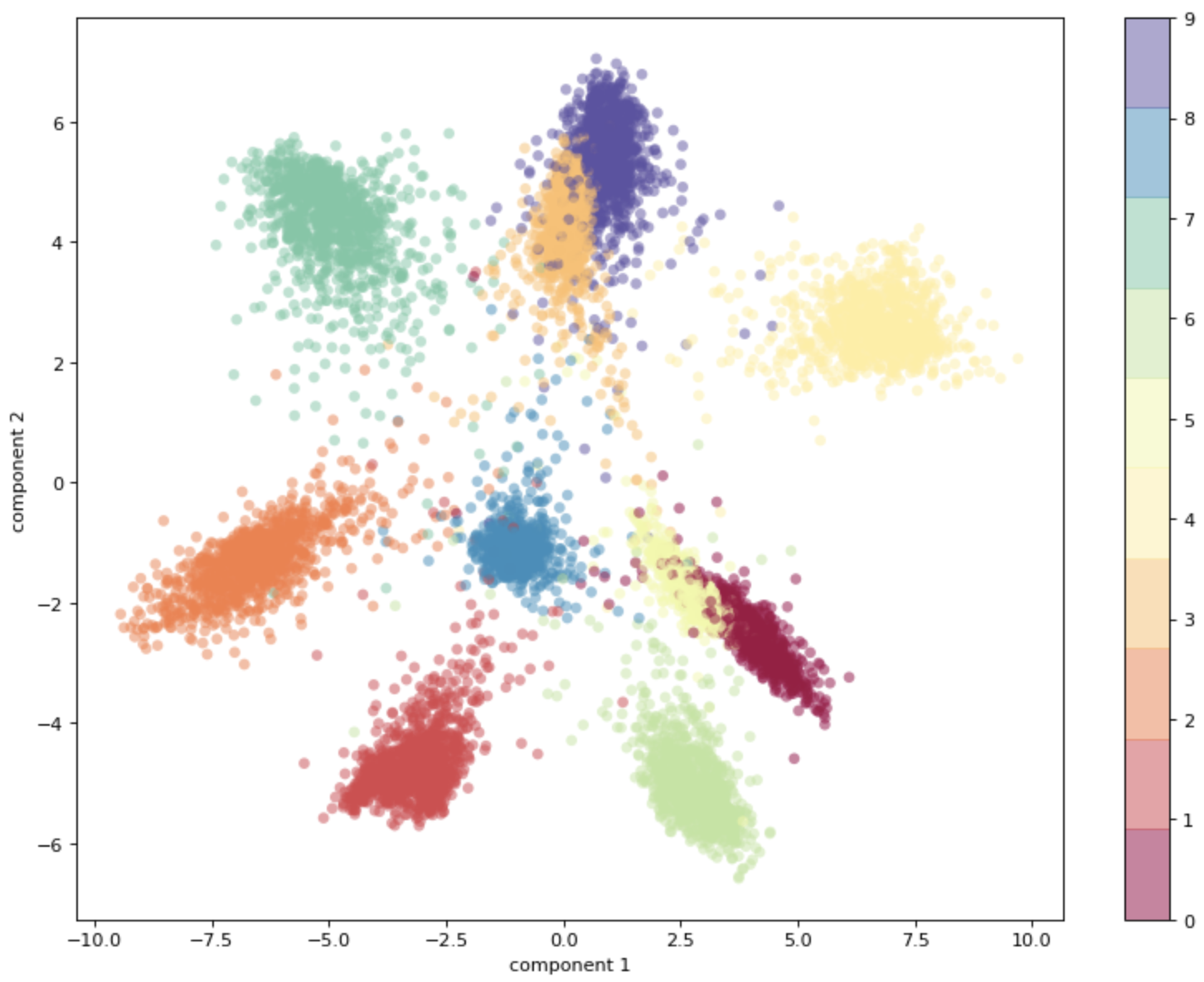}}
\caption{Supervised (pure) side information - Latent space visualization of the test images.}
\label{fig:ls:supervised}
\end{figure}

In the first case, termed "Pure", the side information for the model are the labels themselves. This scenario can be seen as supervised learning and it is one extreme end of the spectrum between supervised and unsupervised settings. Figure~\ref{fig:ls:supervised} depicts the results of this case. As expected, the model was able to create an excellent separation between clusters. Because this case is supervised we can use the labels to obtain an accuracy score. The accuracy achieved was 0.989.

\begin{figure}[htbp]
\centerline{\includegraphics[scale=0.15]{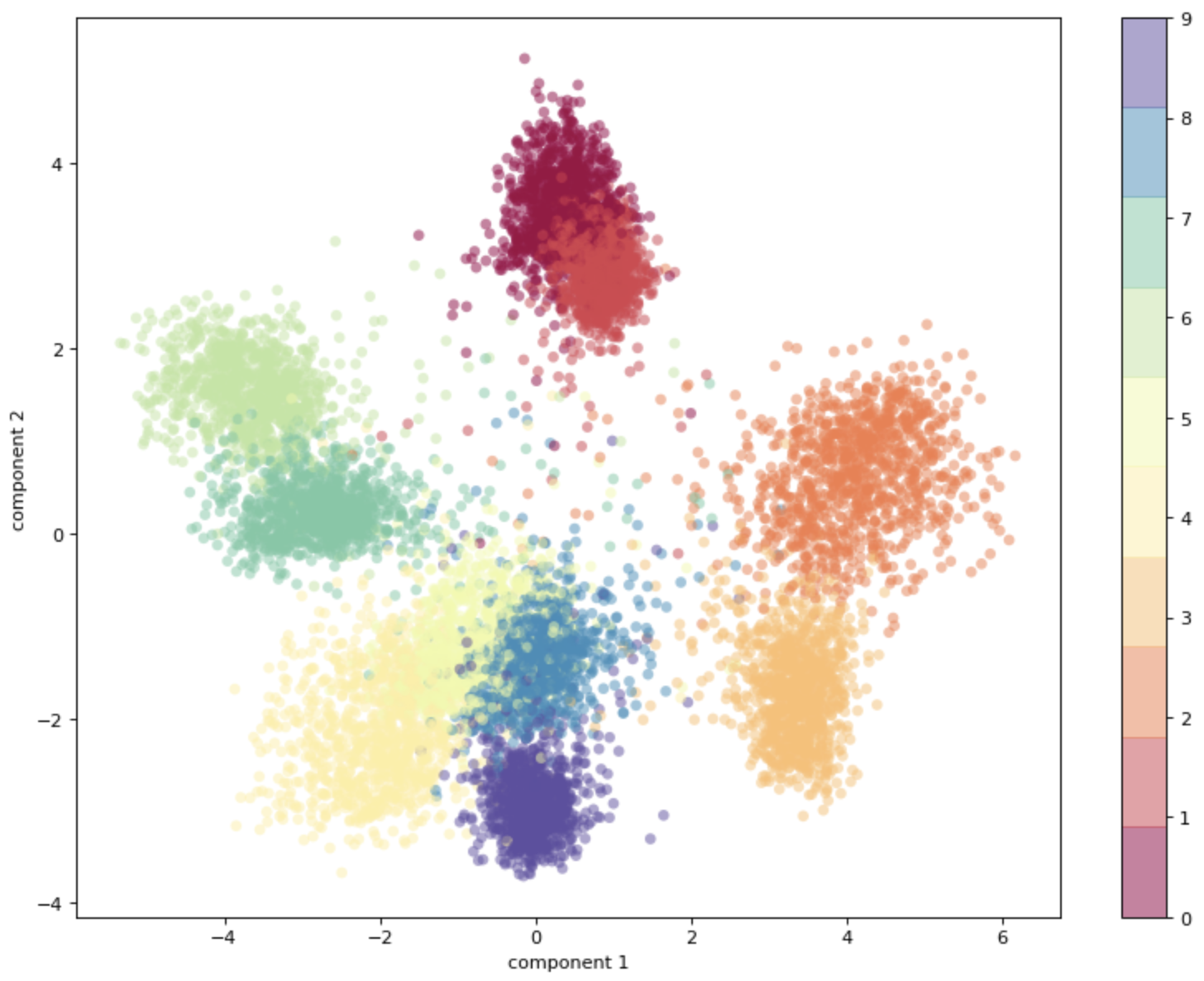}}
\caption{Weak supervision as side information - Latent space visualization of the test images.}
\label{fig:ls:pairs}
\end{figure}

In another case the side information provided to the model are modified labels such that every two label classes are mapped to one side information class. For example, the labels 0 and 1 are mapped to 0, labels 2 and 3 are mapped to 1 and so on. This case simulates a (simple) weak supervision setting. It can be seen in figure \ref{fig:ls:pairs} that the side information propagated to the latent space structure and every two labels have been combined to one cluster. Although the labels were combined, the model was able to maintain separation between the labels in each cluster. 

\begin{figure}[htbp]
\centerline{\includegraphics[scale=0.15]{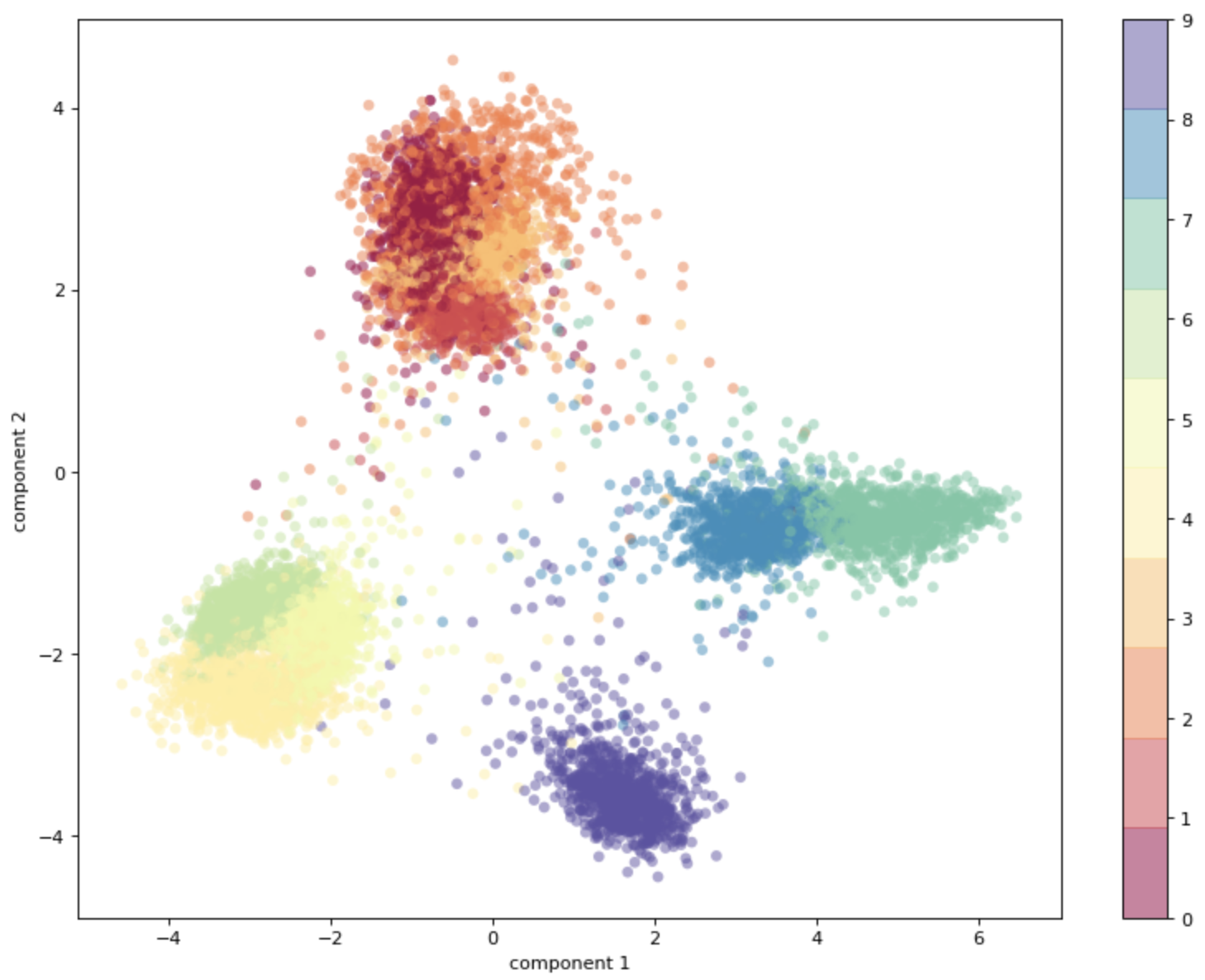}}
\caption{Heterogeneous Weak supervision as side information - Latent space visualization of the test images.}
\label{fig:ls:unbalanced}
\end{figure}

In the last case the labels are mapped in the following way - labels 0,1,2 and 3 are mapped to 0. Labels 4,5 and 6 are mapped to 1. Labels 7 and 8 are mapped to 2 and 9 is mapped to 3. This case simulates a slightly more involved setting where the weak supervision is heterogeneous.
Similar to the previous example, we can see in figure \ref{fig:ls:unbalanced} that the model was able to consider the side information in the latent space while maintaining the internal structure of the instances.


\subsection{Semi-supervised}
\begin{figure}[tbp]
\centerline{\includegraphics[scale=0.4]{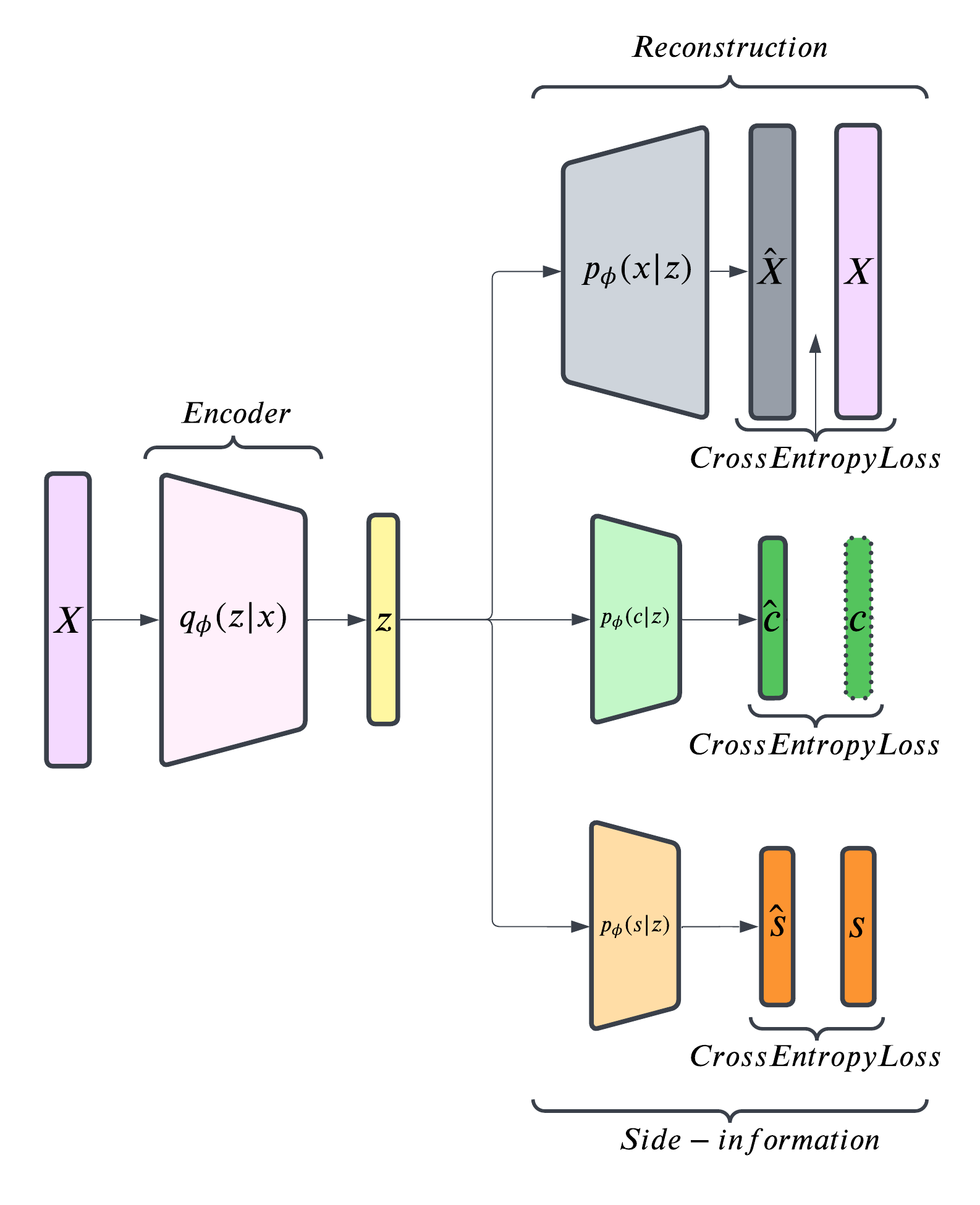}}
\caption{The model's architecture for a semi-supervised use case.}
\label{fig:architecture:semi}
\end{figure}

In this experiment we demonstrate that our model could be used to perform semi-supervised tasks. These tasks contain unlabeled data and also labeled data for a portion of the instances. Using our model for semi-supervised tasks requires a slight addition to the network architecture, depicted in figure~\ref{fig:architecture:semi}. The unlabeled data propagate through the model as before, while for the labeled data, the inferred score $C$ is compared to the true label using cross entropy and that is added to the loss. In order to evaluate the model's performance in semi-supervised tasks, we trained it using 60,000 images, out of which 100 were labeled. Table \ref{tb:semi} shows that our model achieves comparable results on the test-set. Though a few models achieved higher accuracy, they were trained to perform only semi-supervised tasks, unlike our model which is more general and is applicable for a broad range of use cases.

\begin{table}[th]
\begin{center}
\caption{\label{tb:semi}Semi-supervised classification performance (percentage error) for the optimal parameters defined on  MNIST.}

\begin{tabular}{l c c c} 
 \hline
  & & MNIST (100)&  \\ [0.5ex] 
 \hline\hline
  InfoMax \cite{sj:15} & & 33.41 &  \\ 
 \hline
  VAE \cite{fmd:20} & & 14.26 &  \\ 
 \hline
 MV-InfoMax \cite{fmd:20} & & 13.22 &  \\ 
 \hline
   VAE (M1 + M2) \cite{fmd:20} & & 3.33 & \\
  \hline
   IB multiview \cite{fmd:20} & & 3.03 &  \\  
  \hline
   CatGAN \cite{sj:15} & & 1.91 &  \\  
  \hline
   AAE \cite{masj:15} & & 1.90 & \\
 \hline
 Semi-Supervised VIB \cite{VST:20} & & 1.38 & \\
 \hline
  Learning to Score & & 2.58 & \\
 \hline
\end{tabular}
\end{center}
\end{table}

\subsection{Parkinson's Disease Severity Score}
\label{subsec: UPDRS}
UPDRS stands for Unified Parkinson's Disease Rating Scale. It is a standardized assessment tool used by clinicians to measure the severity of Parkinson's disease symptoms and track their progression over time. The UPDRS is derived from assessments of patients' function in several aspects - motor functions, cognitive functions and ability to perform daily activities. 

In this experiment, we tested the performance of our model on a dataset that represents a real-life scenario. The dataset includes subjects’ age, gender and 16 biomedical voice measures. Every subject is labeled with motor UPDRS and total UPDRS. As the motor UPDRS partly contributes to the total UPDRS score we use it as side information. The voice measurements are used as the model's input and the model's predicted scores are matched against 
the total UPDRS scores to evaluate performance (i.e. total UPDRS scores are used only at test time).

\begin{figure}[htbp]
\centerline{\includegraphics[scale=0.32]{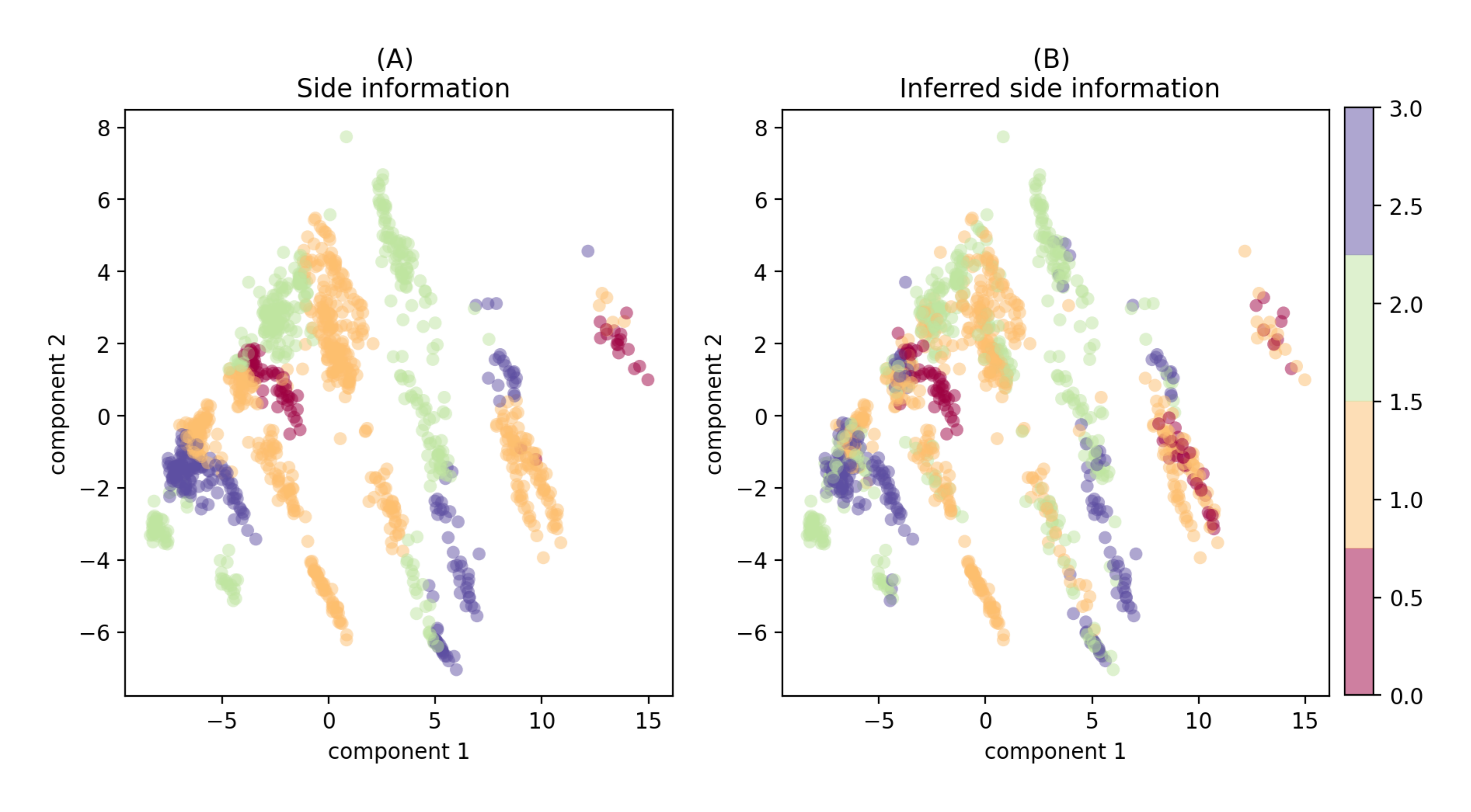}}
\caption{Side information / Inferred side information}
\label{fig:side_info_pred}
\end{figure}

We wanted to examine whether the side information was indeed incorporated in the latent space. Because the side information is used only during training, we can use the side information of the test set to evaluate the degree to which the model learned the side information. Figure \ref{fig:side_info_pred} shows the learned latent space, where the colors of sub-figure A denote the true motor UPDRS scores ($S$), and the colors of sub-figure B denote the inferred motor UPDRS scores ($\hat{S}$).

The main objective of the system is to be able to use the input features and side information to obtain a meaningful score. To examine whether the score produced by the model is in fact meaningful, we examine the correlation with the total UPDRS. As expected, we found a positive correlation of $R=0.58$ ($p < 10^{-107}$) and an accuracy of 0.78, indicating that the model was able to recover discriminative information from the latent representation.  

\ignore{
\subsection{Student Performance Data Set: Mathematics and Portuguese Language Scores}

In this phase of our analysis, we shift our focus towards an educational dataset - the Student Performance Data Set~\cite{c:14}. The dataset represents student achievement of two Portuguese schools. The dataset includes a variety of factors such as student's age, gender, family background, study time, past failures among other features.
}

\subsection{Student Performance Score}

The Student Performance dataset~\cite{c:14} represents  achievement in two Portuguese schools. The dataset includes a variety of factors such as student's age, gender, family background, study time, past failures among other features.

Two separate scores are given in this dataset: a Mathematics score and a Portuguese language score. For this experiment, we use the Mathematics score as the target label while incorporating the Portuguese language score as side information. This setup allows us to evaluate whether students' performance in Portuguese language might give additional context to their Mathematics scores, potentially providing a more comprehensive understanding of their overall academic proficiency.

\begin{figure}[htbp]
\centerline{\includegraphics[scale=0.32]{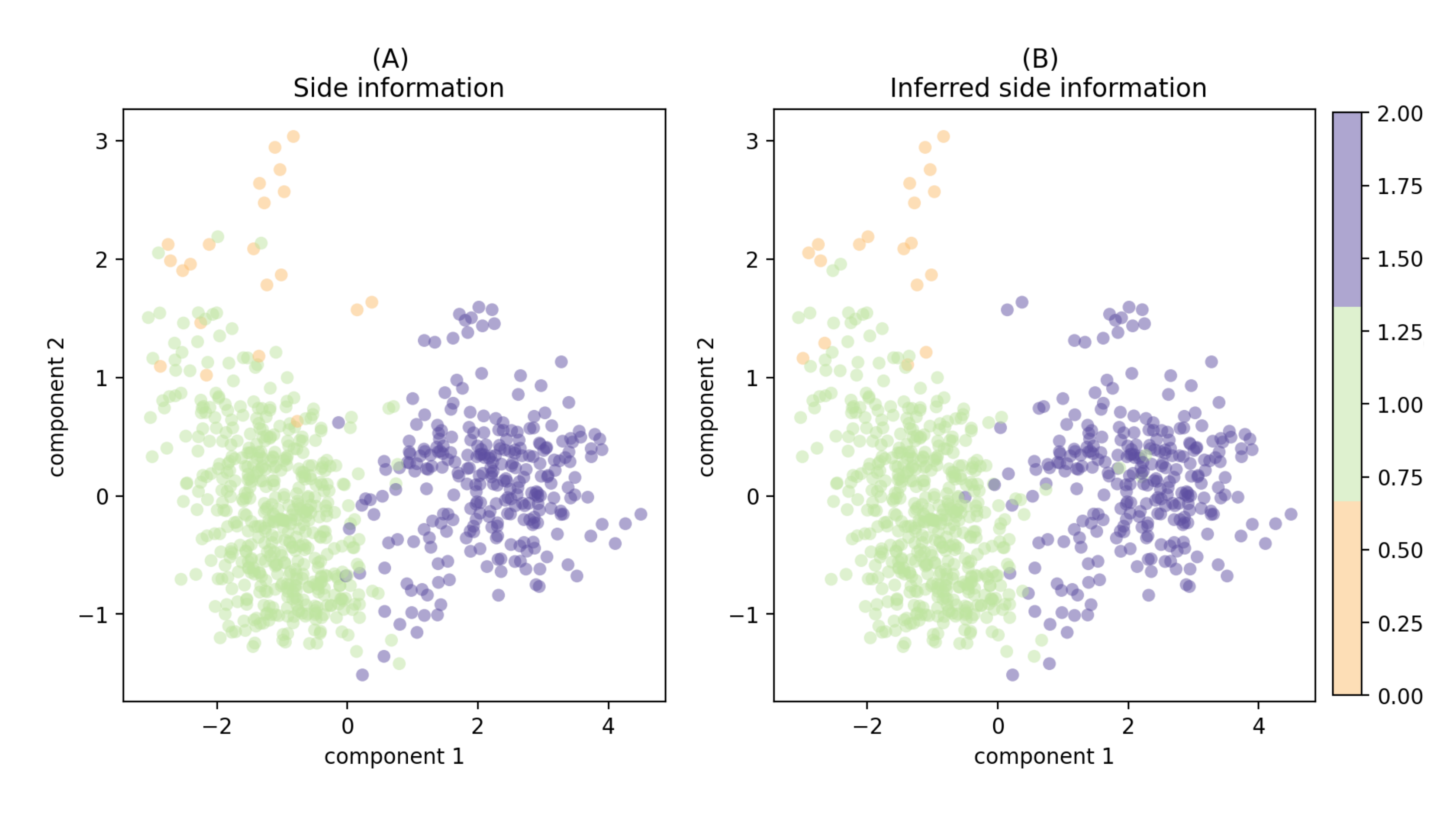}}
\caption{Side information / Inferred side information}
\label{fig:side_info_pred_edu}
\end{figure}

As with the previous experiment, the side information was only utilized during training. Subsequently, we assess its impact by analyzing the test set's side information and its integration within the learned latent space. Figure \ref{fig:side_info_pred_edu} depicts this latent space. The color gradient in sub-figure A signifies the actual Portuguese language scores ($S$), while the gradient in sub-figure B denotes the projected Portuguese language scores ($\hat{S}$).

The primary aim of this model is to effectively leverage the input features and side information to generate an informative score. To validate the meaningfulness of the model's generated scores, we utilized the same technique that was used to validate the UPDRS score (c.f. Section~\ref{subsec: UPDRS}) and examined the correlation with the Mathematics scores. The result displayed a positive correlation of $R=0.49$ ($p < 10^{-300}$) and an accuracy of 0.73. These findings suggest that the model was successful in extracting useful insights from the latent representation, thereby demonstrating that student performance in Portuguese language can provide significant contextual understanding when analyzing their Mathematics scores. This presents an exciting opportunity to implement multidimensional evaluations in educational settings to achieve a well-rounded appraisal of student performance.

\section{Summary}
\label{sec:summary}

In this paper we introduce a novel scoring model, that can handle a wide range of scenarios including absence of labeled data or a well-defined scoring scheme. Additionally, it utilizes supplementary information related to the score, in cases where such information exists. We formalize it as a constraint representation learning. 
Our proposed model is assembled from three semantic components: representation learning, side information and metric learning.
Essentially, we use a generative modeling approach to construct a latent space but strive to form a representation that is as discriminative as possible. We present a comprehensive examination of our proposed model and its capabilities, highlighting the model’s potential for real-world applications and its ability to handle various scenarios and real-life data.

Overall, our empirical results demonstrate the effectiveness of our model in scoring tasks. We believe that these results can contribute to the advancement of scoring models and have practical implications for real-world applications.

In this work we mainly used generative models as the lead modelling approach. It may be advantageous to introduce adversarial learning to this framework, in order to sharpen its  discriminative power.
Additionally, it will be interesting to apply 
the scoring model to more data sets of real-life settings from various fields. 
These directions are left to future research efforts.

%


\bibliographystyle{IEEEtran}
\bibliography{ref}

\vspace{12pt}

\end{document}